%% file: egpaper_arxiv.tex
\documentclass[10pt,twocolumn,letterpaper]{article}

\usepackage[accsupp]{axessibility}  
\usepackage{iccv}
\usepackage{times}
\usepackage{epsfig}
\usepackage{graphicx}
\usepackage{amsmath}
\usepackage{amssymb}

\usepackage{import}
\usepackage{float}
\usepackage{multirow, multicol}
\usepackage[linesnumbered,ruled,vlined]{algorithm2e}
\usepackage{paralist, tabularx}
\usepackage{arydshln}
\usepackage{marvosym}
\usepackage{xcolor} 
\definecolor{darkgreen}{rgb}{0.04,0.63,0.07} 

\usepackage[pagebackref=true,breaklinks=true,letterpaper=true,colorlinks,bookmarks=false]{hyperref}

\iccvfinalcopy 

\def\iccvPaperID{10220} 

\ificcvfinal\fi

\begin{document}

\title{HiLo: Exploiting High Low Frequency Relations for \\Unbiased Panoptic Scene Graph Generation}

\author{Zijian Zhou\\
\small Department of Informatics\\
\small King's College London\\
{\tt\small zijian.zhou@kcl.ac.uk}
\and
Miaojing Shi\textsuperscript{\Letter}\\
\small College of Electronic and Information Engineering \\
\small Tongji University\\
{\tt\small mshi@tongji.edu.cn}
\and
Holger Caesar\\
\small Intelligent Vehicles Lab\\
\small Delft University of Technology\\
{\tt\small h.caesar@tudelft.nl}
}

\maketitle
\ificcvfinal\thispagestyle{empty}\fi

\begin{abstract}
\footnotetext{\textsuperscript{\Letter} Corresponding author.}
\import{}{0_abstract.tex}
\end{abstract}

\section{Introduction}
\import{}{1_introduction.tex}

\section{Related Work}
\import{}{2_related_work.tex}

\section{Method}
\import{}{3_method.tex}

\section{Experiments}
\import{}{4_experiments.tex}

\vspace{-2mm}
\section{Conclusion}
\import{}{5_conclusion.tex}

\vspace{-2mm}
\section*{Acknowledgment}
\vspace{-2mm}
The authors would like to thank Prof. Tomasz Radzik for helpful discussions.
Computing resources provided by King’s Computational Research, Engineering and Technology Environment (CREATE).
This work was supported by the European Union’s Horizon 2020 FET Proactive Program under Agreement 101017857 and Fundamental Research Funds for the Central Universities.

\newpage
{\small
\bibliographystyle{ieee_fullname}
\bibliography{egbib}
}

\newpage

\section*{Supplementary material for HiLo}
\import{}{6_appendix.tex}

\end{document}

%% file: 0_abstract.tex
Panoptic Scene Graph generation (PSG) is a recently proposed task in image scene understanding that aims to segment the image and extract triplets of subjects, objects and their relations to build a scene graph.
This task is particularly challenging for two reasons. 
First, it suffers from a long-tail problem in its relation categories, making naive biased methods more inclined to high-frequency relations.
Existing unbiased methods tackle the long-tail problem by data/loss rebalancing to favor low-frequency relations.
Second, a subject-object pair can have two or more semantically overlapping relations.
While existing methods favor one over the other, our proposed HiLo framework lets different network branches specialize on low and high frequency relations, enforce their consistency and fuse the results.
To the best of our knowledge we are the first to propose an explicitly unbiased PSG method.
In extensive experiments we show that our HiLo framework achieves state-of-the-art results on the PSG task. We also apply our method to the Scene Graph Generation task that predicts boxes instead of masks and see improvements over all baseline methods.
Code is available at \url{https://github.com/franciszzj/HiLo}.

%% file: 1_introduction.tex
 Scene Graph Generation (SGG)~\cite{lu2016visual} is a crucial task in image scene understanding that extracts triplets in the form of subjects, objects and their relations to build a scene graph. 
 Subjects and objects are represented with bounding boxes.
 Since this task links vision and text, it holds great potential for a variety of applications, including visual question answering \cite{hildebrandt2020scene}, image captioning \cite{gao2018image, chen2020say}, image retrieval \cite{johnson2015image, schuster2015generating, qi2017online} and visual reasoning \cite{aditya2018image, shi2019explainable}.

 Recently a novel variant of SGG was proposed, which is Panoptic Scene Graph generation (PSG) \cite{yang2022panoptic}.
 Subjects and objects are represented with panoptic segmentation~\cite{kirillov2019panoptic} masks. Unless stated otherwise, in this work we focus on PSG, since it is pixel-level accurate and also covers background classes and their relations with foreground objects.

\begin{figure}
\begin{center}
\includegraphics[width=\columnwidth]{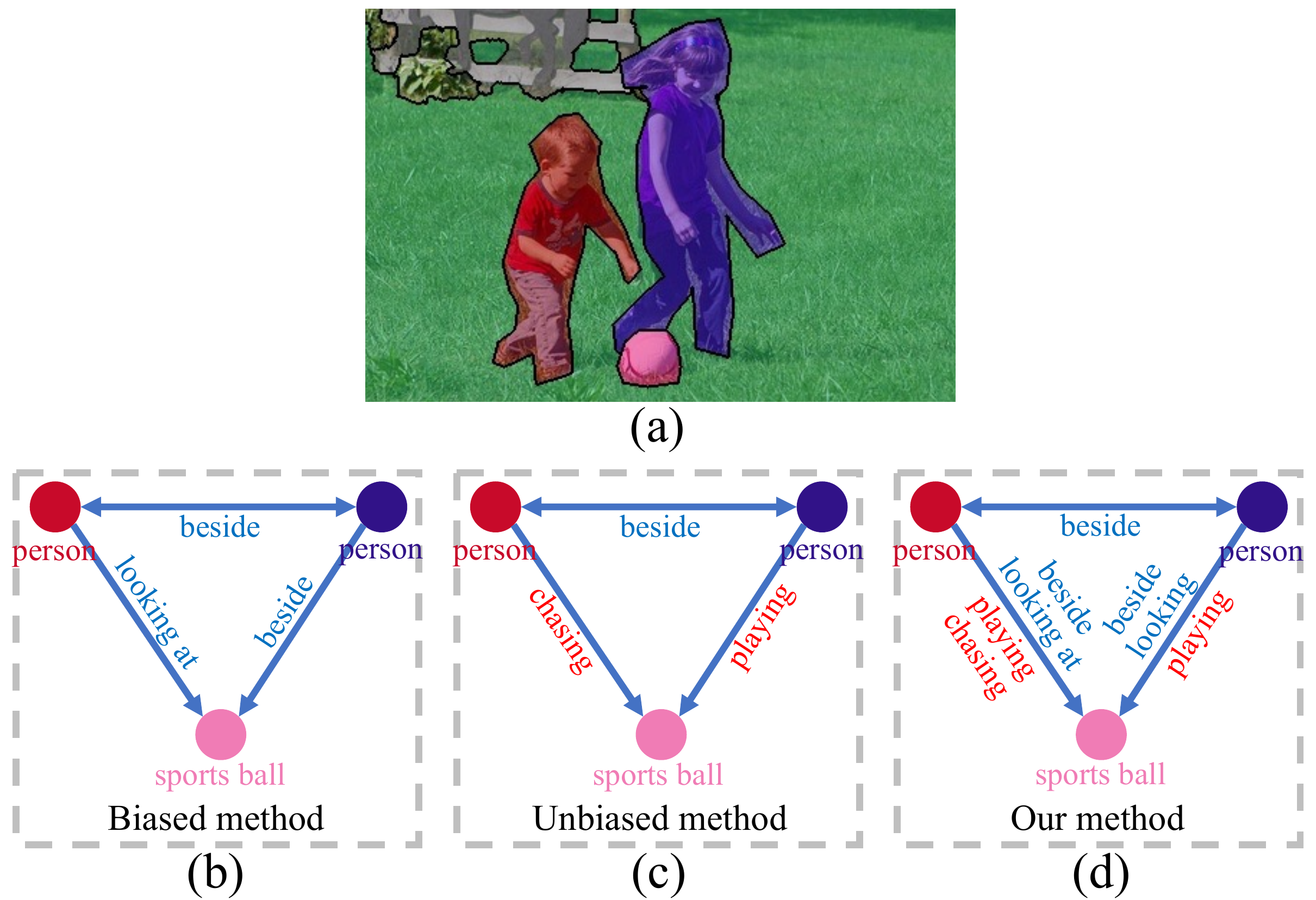}
\end{center}
\vspace{-0mm}
\caption{
An example of the PSG task. We compare the predicted object relations of different approaches. 
a) An image and its panoptic segmentation. Two persons and a sports ball are shown in different colors.
b) A biased method predicts mostly high frequency relations.
c) An unbiased method predicts mostly low frequency relations.
d) Our method predicts both low and high frequency relations, as well as more relations in total.
}
\vspace{-4mm}
\label{figure:motivation}
\end{figure}

The performance of the PSG model is affected by a \emph{long-tail problem} in its relation categories.
For instance, relations such as \textit{over}, \textit{in front of} and \textit{holding} occur tens of thousands of times in the PSG dataset~\cite{yang2022panoptic}, while others like \textit{swinging} and \textit{kissing} occur only a few dozen times.
This severe class imbalance in the relation categories can lead to model predictions that are more inclined to high-frequency relations, which poses significant challenges to the application of panoptic scene graphs in real-world scenarios.

Previous methods \cite{chiou2021recovering, desai2021learning, li2021bipartite, deng2022hierarchical} have often treated the long-tail problem of the PSG task as equivalent to the long-tail problem in object-centric tasks such as classification \cite{cui2019class, cao2019learning, hong2021disentangling} or semantic segmentation \cite{cui2022region, alexandridis2022long}.
Consequently, these methods have employed re-balancing techniques to address class imbalance, either through re-sampling the data \cite{li2021bipartite} or by using a class-balanced loss~\cite{kang2023skew} that assigns different weights to different relation categories.

In contrast, in the relation-centric PSG task, a subject-object pair can have multiple relations that exhibit \textit{relational semantic overlap}, such as being partially or fully overlapping.
For example, in Fig.~\ref{figure:motivation}, there are multiple relations between the boy and the sports ball, such as \textit{beside}, \textit{looking at}, \textit{playing} and \textit{chasing}.
For regular biased models \cite{xu2017scene, zellers2018neural, tang2019learning, lin2020gps}, the results are dominated by high-frequency relations (\textit{beside}, \textit{looking at}).
For specifically unbiased models \cite{zhang2022fine, yang2022panoptic}, the results are dominated by low-frequency relations (\textit{playing}, \textit{chasing}).
However, since the low frequency relations can be more specific (e.g. \textit{on} and \textit{standing on}) or only partially overlapping with high frequency relations (e.g. \textit{looking at} and \textit{chasing}), it is crucial to include both to fully understand the image.
We found that relational semantic overlap occurs in large numbers in the PSG dataset~\cite{yang2022panoptic} and that current methods do not effectively address it.
This is reflected in the increase in the category-averaged \textit{mean recall} metric of unbiased methods, at the cost of the decrease in global \textit{recall} (see Sec.~\ref{sec:experiments:main_results}).

To address the long-tail problem of scene graphs under relational semantic overlap, we introduce the HiLo framework. This framework simultaneously learns the high and low frequency relations in different network branches and unifies their strengths with the help of two novel consistency loss functions.
We apply our framework on top of a novel baseline. This baseline uses a recent transformer-based approach~\cite{cheng2022masked} for panoptic segmentation and adapts triplet queries~\cite{yang2022panoptic} and masked attention~\cite{cheng2022masked} for the PSG task.
In summary, we make the following contributions:
\begin{itemize}
    \vspace{-2mm}
    \item We identify the long-tail problem with relational semantic overlap in the PSG task and propose the HiLo framework to address this problem. The framework is general and can be applied to any PSG method.

    \vspace{-2mm}
    \item We propose a powerful and efficient one-stage end-to-end baseline. This baseline enhances the interaction between mask and relation prediction in the transformer decoder layer.

    \vspace{-2mm}
    \item We conduct extensive experiments to demonstrate the effectiveness of our framework and baseline.
    Our results outperform the state-of-the-art in both recall and mean recall on the PSG dataset and show systematic improvements on the VG dataset.
\end{itemize}

%% file: 2_related_work.tex
\begin{figure*}[h!]
\begin{center}
\includegraphics[width=\linewidth]{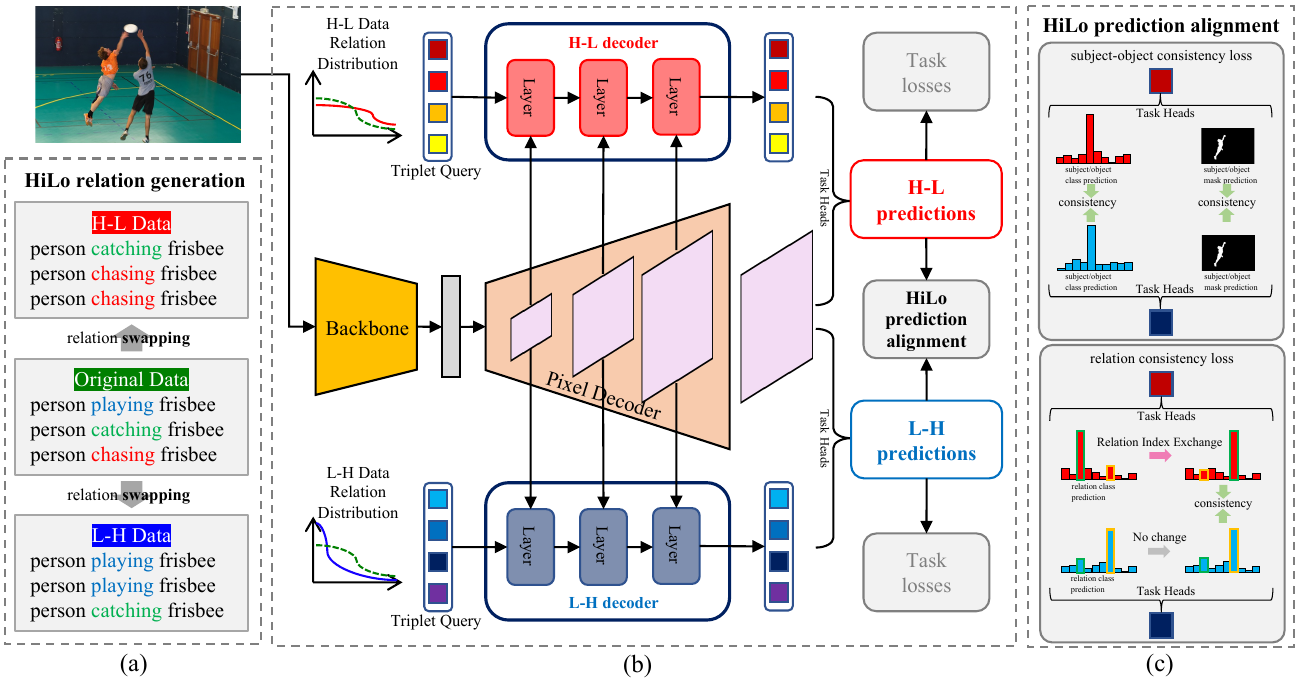}
\end{center}
\vspace{-2mm}
\caption{An overview of our HiLo framework with HiLo baseline.
a) HiLo relation swapping module swaps the multiple relations in the subject-object pair to obtain H-L Data and L-H Data respectively.
b) Input data into our HiLo framework with HiLo baseline model, there are two branches, namely H-L decoder and L-H decoder, which learn H-L Data and L-H Data respectively.
c) In addition to task losses for PSG, we propose HiLo prediction alignment, which includes subject-object consistency loss and relation consistency loss, so that the parallel branch can be better optimized.
}
\vspace{-4mm}
\label{figure:hilofex_baseline}
\end{figure*}

\subsection{Scene Graph Generation}
The Scene Graph Generation (SGG) \cite{lu2016visual} task plays a crucial role in connecting vision and language, and has received widespread attention in the computer vision community.
Many methods have been proposed to improve the performance of SGG, which can be classified into three categories \cite{zhu2022scene}.
The first is to introduce multi-modal information, such as appearance \cite{sadeghi2011recognition}, space \cite{zhu2017visual}, depth \cite{sharifzadeh2021improving}, and segmentation \cite{khandelwal2021segmentation}.
The second is to introduce prior information and commonsense knowledge, such as statistical \cite{baier2017improving, dai2017detecting, zellers2018neural, chen2019knowledge} and language prior knowledge \cite{lu2016visual, liao2019natural, zhang2019large, hwang2018tensorize, dupty2020visual}.
The third category involves designing different model structures, such as message passing \cite{li2017vip, dai2017detecting, li2017scene, zellers2018neural, gu2019scene, hu2022neural}, attention mechanisms \cite{zheng2019visual, qi2019attentive}, tree structures and visual translation \cite{zhang2017visual, hung2020contextual}. However, most of these methods are two-stage methods that cannot learn scene graphs end-to-end.
In contrast, to improve the learning ability of SGG models, several methods based on transformers have been proposed, including SGTR \cite{li2022sgtr}, RelationFormer \cite{shit2022relationformer} and RelTR \cite{cong2023reltr}. These are end-to-end trainable in a single stage.

\subsection{Unbiased Scene Graph Generation}
Solving the long-tail problem in the SGG task has attracted considerable attention from researchers, and several unbiased methods \cite{tang2020unbiased, yu2020cogtree, chiou2021recovering, desai2021learning, guo2021general, li2021bipartite, dong2022stacked, goel2022not, li2022devil, li2022ppdl, deng2022hierarchical, zhang2022fine} have been proposed.
These methods typically improve the model from the perspective of data re-sampling~\cite{li2021bipartite} or a class-balanced loss~\cite{kang2023skew}.
BGNN \cite{li2021bipartite} uses a two-layer re-sampling strategy to provide a more balanced data distribution during training, while CogTree \cite{yu2020cogtree} proposal exploits the semantic relation between different predicate classes to design a novel CogTree loss.
HML \cite{deng2022hierarchical} improves the model's ability to solve long-tail problems by designing a staged training process, and IETrans~\cite{zhang2022fine} proposes an internal and external transfer method to transfer high frequency relations to low frequency relations and recover missing relations to train the unbiased model.
Dong \etal~\cite{dong2022stacked} propose to group relations by their frequency and train specialized relation encoders for each group. 
While these methods are able to mitigate the long-tail problem, they do not address relational semantic overlap.

\subsection{Panoptic Scene Graph Generation}
Contrary to SGG, the PSG task~\cite{yang2022panoptic} uses panoptic segmentation masks instead of bounding boxes to represent objects, resulting in a more comprehensive scene graph.
PSGTR \cite{yang2022panoptic}, an end-to-end method based on the DETR structure, was proposed to construct a transformer-based PSG model.
PSGFormer~\cite{yang2022panoptic} further improved on PSGTR by introducing Object \& Relation Query Learning Blocks and Query Matching Blocks. Their strong performance on most relation classes indicates that their method is implicitly unbiased. In contrast, our approach is the first to explicitly bias a PSG model, creating separate branches for low and high frequency relations, which are then fused together.

Since the scene graph in PSG is built upon the subjects, objects and their relations, strong panoptic segmentation~\cite{kirillov2019panoptic} is crucial for PSG.
Several DETR-based \cite{carion2020end} methods such as Deformable-DETR \cite{zhu2020deformable}, Segmenter \cite{strudel2021segmenter}, MaskFormer \cite{cheng2021per} and Mask2Former \cite{cheng2022masked} have recently pushed the envelope in panoptic segmentation.
These methods have introduced deformable transformer encoders and decoders, as well as pixel decoders and multi-scale information to improve model performance and speed up convergence.
Our proposed baseline builds upon these techniques to further enhance the PSG performance and speed up model convergence.

%% file: 3_method.tex
\subsection{Problem Setting}
The Panoptic Scene Graph generation (PSG) task aims to generate a panoptic scene graph $\mathcal G$ for a given image $\mathcal{I} \in \mathbb{R}^{H \times W \times 3}$, where $\mathcal G$ contains an object set $\mathcal O$ and a relational triplet set $\mathcal T$, 
denoted by $\mathcal{G} = (\mathcal{O}, \mathcal{T})$.
For the $i$-th object $o_i$ in $\mathcal O = \{o_i\}_{i=1}^{N}$, we use 
 $m_i$ to represent the object's mask and   $c_i$  the object's category, \ie $o_i= (m_i, c_i)$. For the $j$-th triplet  in $\mathcal T = \{t_j\}_{j=1}^{M}$, we use $s_j$  to represent the subject, $o_j$ the object, and $r_j$ their relation, \ie $t_j = (s_j, o_j, r_j)$. There are in total $C$ object categories and $R$ relation categories. For the $k$-th relation, we denote its frequency in the training set as  $f_k$. 

\subsection{HiLo Baseline}
\label{sec:baseline}
High-quality panoptic segmentation is crucial for achieving good PSG performance.
We build our method upon the latest advances in DETR-based panoptic segmentation, Mask2Former \cite{cheng2022masked}.
Below we present its structure, as well as our proposed modifications for the PSG task.  

\noindent \textbf{Mask2Former}.
This method comprises three key parts: A backbone (CNN-based or transformer-based) followed by a pixel decoder, a transformer module with a transformer decoder and a task-specific module with different task heads. 
Specifically, the backbone takes an input image $\mathcal I$ and generates an image feature $F$.
The pixel decoder then gradually upsamples $F$ to produce multi-scale features $\mathcal{\widetilde F} = \{\widetilde F_i\}_{i=1}^4$.
The transformer decoder takes a set of queries $\mathcal {Q}$ of size $N$ and multi-scale features $\mathcal{\widetilde F}$ as input and outputs a set of mask features $\mathcal X$ of the same size with $\mathcal{Q}$.

On top of the transformer decoder, there are two task heads including a linear classifier that predicts the class probability for each mask; and a multi-layer perceptron (MLP) that uses the mask features $\mathcal X$ to  generate the mask embedding $\mathcal{E}$.
The mask prediction is obtained by taking the dot product of the mask embedding $\mathcal{E}$ with the scale feature of the highest resolution in the multi-scale features $\mathcal{\widetilde F}$.

\noindent \textbf{Triplet queries}.
The original query in Mask2Former is to predict the object. In order to predict both subject, object and their relation, we develop the triplet queries $\mathcal{Q}^{t}$ inspired by \cite{yang2022panoptic} into our baseline.
Each query predicts a triplet that includes subject, object, and relation.
Accordingly, our task heads comprise three linear classifiers. Each classifier is responsible for predicting the class probability for subject, object or relation, respectively. 
Moreover, we devise two MLPs to generate mask embeddings for the subject and object, denoted by $\mathcal{E}^s$ and $\mathcal{E}^o$. We use them to compute the dot product with the scale feature of the highest resolution in multi-scale feature $\mathcal{\widetilde F}$ and obtain the mask prediction for the subject and object, respectively.

\noindent \textbf{Masked relation attention}.
\cite{cheng2022masked} proposes a transformer decoder variant with masked attention, which extracts spatial features by adding the predicted mask of the object from the previous decoder layer's mask prediction.
It makes the model focus on the object-related area in the feature map.
To adapt this scheme to the PSG task, for each relation, we extend the masked attention to take the union of the binary masks of the subject and object as input, which represents the pixels corresponding to the relation.

\noindent \textbf{Network training}.
We adopt the same losses as PSGTR~\cite{yang2022panoptic}, including cross-entropy loss $\mathcal{L}_{so\_cls}$ for object classification of subject-object pairs and a combination $\mathcal{L}_{so\_mask}$ of focal loss \cite{lin2017focal} and dice loss \cite{sudre2017generalised} to jointly supervise mask learning.
To supervise the relations, we use the cross-entropy loss $\mathcal{L}_{rel\_cls}$.
The baseline loss with (${\lambda}_{1} = 1$, ${\lambda}_{2} = 1$, ${\lambda}_{3} = 4$) is thus:
\vspace{-2mm}
\begin{align}
\mathcal{L}_{baseline} = {\lambda}_{1} \cdot \mathcal{L}_{so\_cls} + {\lambda}_{2} \cdot \mathcal{L}_{so\_mask} + {\lambda}_{3} \cdot \mathcal{L}_{rel\_cls}
\end{align}
\vspace{-2mm}

\subsection{HiLo Framework}
\label{sec:framework}

The key insight of our HiLo is to build a model that can take into account both high frequency and low frequency relations, and effectively improve the performance on low frequency relations without degrading the performance on high frequency relations.

\subsubsection{HiLo relation generation}
\label{sec:framework:relation_generation}
In the PSG task, multiple relations can be used to describe the connection between a subject and an object from different perspectives, such as \emph{spatial} relations, \emph{actions} and \emph{prepositions}. Since the subject, object and their relative position in the image are fixed, all these relations share the same visual information.
This is reflected in the PSG dataset~\cite{yang2022panoptic} where many subject-object pairs are annotated with multiple relation labels. These relation labels are of different frequencies in the dataset. In this section we introduce the HiLo relation generation module to prepare two sets of training data, biasing towards high and low-frequency relations respectively.

\noindent \textbf{Relation augmentation}. 
Similar to~\cite{goel2022not}, we observe that there are many missing relation annotations in the PSG dataset~\cite{yang2022panoptic}.
To add the missing relation annotations, we design a relation augmentation scheme that is inspired by IETrans \cite{zhang2022fine}, which converts high-frequency relations to low-frequency relations and adds a relation to a subject-object pair that has no relation.
We adapt it by first training our baseline as a biased model using the original annotated data. For every subject-object pair in the training set, we use this model to predict the relation scores for all predefined relation categories: 
\begin{compactitem}
    \item If this subject-object pair has annotated relation labels, we pick the one with the highest predicted score and use the score as a threshold. For other relations (except for already labeled ones) whose predicted scores are greater than the threshold, we add them as the relation labels of this subject-object pair.
    \item If this subject-object pair has no annotated relation labels, we use the predicted score of the \emph{no-relation} class as a threshold. For other relations whose predicted scores are greater than this threshold we add them as the relation labels of this subject-object pair.
\end{compactitem}
This operation allows us to significantly augment the relation labels for the subject-object pairs in the training data, which is especially relevant for pairs with zero or no annotated relations. 

\noindent \textbf{Relation swapping}.
We swap the relation labels for each subject-object pair. 
Specifically, given a subject-object pair $(s, o)$ with $K$ relation labels, we have $K$ triplets $(s, o, r_1),...,(s, o, r_K)$, sorted by their relation frequency in descending order, $f_1 >... > f_K$. We denote by H-L and L-H the swapping of high-frequency relations with low-frequency relations and vice versa. This creates two sets of data: 

\emph{H-L Data.}
Given a triplet ($s, o, r_k$), we replace its relation label $r_k$ with that of the next triplet $r_{k+1}$ with lower frequency. We sequentially process all triplets from ($s, o, r_1$) until ($s, o, r_{K-1}$), keeping the last triplet unchanged. 

\emph{L-H Data.}
Given a triplet ($s, o, r_k$), we replace its relation label $r_k$ with that of the previous triplet $r_{k-1}$ with higher frequency. 
 We sequentially process all triplets from ($s, o, r_K$) until ($s, o, r_2$), keeping the first triplet unchanged.
Hence, we obtain two new sets of triplets, denoted by $\mathcal T^\text{H-L}$ and $\mathcal T^\text{L-H}$.  We devise two parallel decoders from the shared encoder of our backbone. They are learned with the H-L and L-H data, respectively.
The H-L and L-H decoders favour the predictions for low and high-frequency relations, respectively. Despite their difference, for the corresponding triplet query in the two decoders, their predictions are highly correlated: on one hand, the subject and object predictions should be the same; on the other hand, the distribution of relation predictions should be overlapping.  Below we first build the query correspondence in the two decoders and then introduce the HiLo subject-object and relation consistency loss to align the predictions from two the decoders. 

\subsubsection{HiLo prediction alignment}
\label{sec:framework:prediction_alignment}
Training two different relation distributions simultaneously confuse the model.
To make the model differentiate between the two branches, we built a HiLo prediction alignment module, including triplet query correspondence, subject-object consistency loss and relation consistency loss respectively.

\noindent \textbf{Triplet query correspondence}.
In order to construct the subject-object consistency loss and relation consistency loss, we first need to construct the correspondence between the triplet queries in the H-L and L-H decoders.
The same query index in the two decoders is not naturally matched.  In order to find the query correspondence, we need to rely on the ground truth assignment: we use Hungarian matching to assign the triplet label to the corresponding triplet query, and record the relation label index corresponding to the triplet query.
This label index allows us to construct the correspondence between the triplet queries in both decoders.
We calculate the consistency loss for the triplet query prediction with the same relation label index in the two decoders.

\noindent \textbf{Subject-object consistency loss}.
Having the corresponding predictions from the two decoders, both their subjects and objects have the same ground truth and should be equal. We propose a subject-object consistency loss $\mathcal L_{obj}$ to minimize the mean squared error (MSE) of the corresponding predictions from the two decoders. 

\vspace{-4mm}
\begin{align}
\mathcal L_{obj} &= \mathcal L_{cls} + \mathcal L_{mask} \\
\mathcal L_{cls} &= \lVert softmax(p_{c}^\text{H-L}) - softmax(p_{c}^\text{L-H}) \rVert^2 \\
\mathcal L_{mask} &= \lVert sigmoid(p_{m}^\text{H-L}) - sigmoid(p_{m}^\text{L-H}) \rVert^2
\end{align}
Here $\mathcal L_{cls}$ and $\mathcal L_{mask}$ represent the MSE losses for class prediction and mask prediction, respectively.

$p_{c}^\text{H-L}$ and $p_{c}^\text{L-H}$ are the class prediction logits from the H-L and L-H decoders. After a softmax, they are $C$-dimensional probability vectors. After a sigmoid, $p_{m}^{H-L}$ and $p_{m}^{L-H}$ are the mask prediction logits for the H-L and L-H decoders.  

\noindent \textbf{Relation consistency loss}.
Given a pair of subject and object, 
we have previously swapped the high-low frequency relation labels to create data for the H-L and L-H decoders. For H-L, the prediction of the low-frequency relation logit is of high value; while for L-H, the prediction of the high-frequency relation logit is of high value. For the predictions on the rest logits, they should be similar, since it is the same pair of subject and object for the two decoders. Based on this observation, we introduce the relation consistency loss.

Specifically, for a pair of subject and object, we use $p_{r}^\text{H-L}$ to denote the predicted relation logits from  the H-L decoder, $p_{r}^\text{L-H}$ the predicted relation logits from the L-H decoder. They are $R$-dimensional probability vectors after softmax. We can map between the distributions of the two vectors by swapping the logit values between the high- and low-frequency relation indices, which is named Relational Index Exchange operation.
We use $\text{RIE}(\cdot)$ to denote this operation.
$\text{RIE}(\cdot)$ includes a stop gradient operation, which creates copies from original predictions of two branches, enabling value exchanges.
For example, for relations $r_k$ and $r_{k+1}$, $\text{RIE}(\cdot)$ exchanges the values between $p_{r,k}$ and $p_{r,{k+1}}$ in $p_r$. We can therefore compute the distance between $p_{r}^\text{H-L}$ and its mapped  counterpart from $p_{r}^\text{L-H}$,  vice versa: 

\vspace{-4mm}
\begin{align}
\begin{split}
\text{Dist}_\text{HiLo} = &\lVert softmax(p_{r}^\text{H-L}) - \mathit{RIE}(softmax(p_{r}^\text{L-H})) \rVert ^ 2 \\
+ &\lVert \mathit{RIE}(softmax(p_{r}^\text{H-L})) - softmax(p_{r}^\text{L-H}) \rVert ^ 2
\end{split}
\end{align}
The relation consistency loss is defined to minimize $\text{Dist}_\text{HiLo}$ with a margin of $m$:
\vspace{-2mm}
\begin{align}
\label{eq:relation_consistency_loss}
\mathcal L_{rel} = max(\text{Dist}_\text{HiLo}- m, 0),
\end{align}
where $m$ is a small constant.  Adding $m$ is due to the fact that the high- and low frequency relations might be only partially semantically overlapping.    

\noindent \textbf{Network training}.
Our subject-object consistency loss and relation consistency loss can seamlessly integrate with the losses of any baseline method to jointly supervise the training of the entire model. Notably, we supervise the output of each transformer decoder layer to ensure effective learning. 
The final loss $\mathcal{L}$ is thus: 
\begin{align}
\mathcal{L} = \mathcal{L}_{baseline} + \mathcal{L}_{obj} + \mathcal{L}_{rel} \;
\end{align}

\begin{table*}[h!]\normalsize
    \centering
    \vspace{-4mm}
    \begin{tabular}{lc|cc|cc|cc}
    \hline
        ~ & ~ & \multicolumn{6}{c}{Scene Graph Detection} \\
        \cline{3-8}
        Method & Backbone & R@20 & mR@20 & R@50 & mR@50 & R@100 & mR@100 \\
        \hline
        IMP \cite{xu2017scene} & R50 & 16.5 & 6.5 & 18.2 & 7.1 & 18.6 & 7.2 \\
        MOTIF \cite{zellers2018neural} & R50 & 20.0 & 9.1 & 21.7 & 9.6 & 22.0 & 9.7 \\
        VCTree \cite{tang2019learning} & R50 & 20.6 & 9.7 & 22.1 & 10.2 & 22.5 & 10.2 \\
        GPSNet \cite{lin2020gps} & R50 & 17.8 & 7.0 & 19.6 & 7.5 & 20.1 & 7.7 \\
        \hline
        PSGTR \cite{yang2022panoptic} & R50 & 28.4 & 16.6 & 34.4 & 20.8 & 36.3 & 22.1 \\
        PSGFormer \cite{yang2022panoptic} & R50 & 18.0 & 14.8 & 19.6 & 17.0 & 20.1 & 17.6 \\
        \textbf{HiLo (ours)} & R50 & 34.1 & 23.7 & 40.7 & 30.3 & 43.0 & 33.1 \\
        \textbf{HiLo (ours)} & Swin-B & 38.5 & 28.3 & 46.2 & 35.3 & 49.6 & 39.1 \\
        \textbf{HiLo (ours)} & Swin-L & \textbf{40.6} & \textbf{29.7} & \textbf{48.7} & \textbf{37.6} & \textbf{51.4} & \textbf{40.9} \\
        \hline
    \end{tabular}
    \vspace{+1mm}
    \caption{Comparison between our HiLo and other methods on the PSG dataset.
    Our method shows superior performance compared to all previous methods.
    }
    \vspace{-2mm}
    \label{table:psg_main_results}
\end{table*}

\subsubsection{HiLo inference fusion}
\label{sec:framework:inference_fusion}
The H-L and L-H decoders favour low-frequency relation prediction and high-frequency relation prediction, respectively. 
To combine the strength of both during inference, we introduce the HiLo inference fusion module.
Specifically, we denote by $\mathcal G^\text{H-L}$ and $\mathcal G^\text{L-H}$ the predicted panoptic scene graphs from H-L and L-H decoders, respectively.
There are $N_{1}$ triplets in $\mathcal G^\text{H-L}$ and $N_{2}$ triplets in $\mathcal G^\text{L-H}$.

\begin{compactitem}
\item  First, we merge the triplets in $\mathcal G^\text{H-L}$ and $\mathcal G^\text{L-H}$ and sort them in the descending order according to their relation scores. We obtain a list of $N_{1} + N_{2}$ triplets.  

\item Second, starting from the first triplet, we de-duplicate the triplet list. For the $i$-th triplet $\mathcal T_i$, we identify its duplicated versions from the $(i+1)$-th triplet until the end of the list.  
Remove any follow-up triplet from the list if it 1) has the same subject, object and relation classes to that in $T_i$ and 2) has a mask IoU greater than a threshold, \eg 0.5, between the predicted subject/object and the corresponding subject/object in $T_i$.

\item Third, after deduplication, for each triplet in the list, we multiply the relation, subject and object scores as an overall score for it. We sort the  triplet list according to this score in descending order to obtain the final panoptic scene graph $\mathcal G^\text{HiLo}$.

\end{compactitem}

%% file: 4_experiments.tex
\subsection{Datasets}

\noindent \textbf{Panoptic Scene Graph Generation (PSG)~\cite{yang2022panoptic}}.
This is the first Panoptic Scene Graph generation dataset. It has a total of 48,749 labeled images including 2,177 test images and 46,572 training images.
The object categories comprise 80 thing classes and 53 stuff classes, which is the same as the COCO \cite{lin2014microsoft} and COCO-Stuff datasets \cite{caesar2018coco}.
The relation categories comprise 56 classes, including positional relations, common object-object relations, common actions, human actions, actions in the traffic scene, actions in the sports scene and interactions between backgrounds~\cite{yang2022panoptic}. 

\noindent \textbf{Visual Genome (VG)} \cite{krishna2017visual}.
VG is a widely used benchmark dataset for Scene Graph Generation.
Following previous work \cite{zellers2018neural, chen2019counterfactual}, we adopt the widely accepted split, VG-150, which contains 150 object categories and 50 relation categories.
The object categories cover a wide range of classes, such as \textit{animals}, \textit{vehicles} and \textit{household items}.
The relation categories include both spatial and semantic classes, such as \textit{on}, \textit{in} and \textit{wearing}.

\subsection{Tasks and Metrics}
Three subtasks have been proposed for the SGG and PSG tasks, which are \emph{Predicate Classification}, \emph{Scene Graph Classification} and \emph{Scene Graph Detection}~\cite{xu2017scene}.
We focus on Scene Graph Detection for both datasets, since it is the most comprehensive and addressed by \cite{yang2022panoptic}. This subtask
requires the model to first localize the objects and then predict the object classes and relations.
Note that Scene Graph Detection in PSG includes the detection on stuff classes, while in SGG it does not.

Following previous work \cite{tang2019learning, yu2020cogtree, yang2022panoptic}, we use Recall@K (R@K) and mean Recall@K (mR@K) as our metrics, where the former metric is dominated by high-frequency relations, while the latter assigns equal weight to all relation classes.

\subsection{Implementation details}
\label{sec:experiments:implementation}
In our experiments, we follow the training strategy of PSGTR \cite{yang2022panoptic}.
We use the AdamW optimizer \cite{loshchilov2017decoupled}, with a learning rate of $1e^{-4}$ and weight decay of $1e^{-4}$, except for the backbone, which is trained with a learning rate of $1e^{-5}$. For initialization, we use Mask2Former \cite{cheng2022masked} pretrained on COCO \cite{lin2014microsoft} to initialize our backbone and pixel decoder.
Following Mask2Former \cite{cheng2022masked}, we use 100 triplet queries for the H-L and L-H decoders respectively.
Additionally, both the H-L and L-H decoders are initialized with Mask2Former's transformer decoder.
To ensure consistent comparison with PSGTR, we adopted the same data augmentation settings.
Our model is trained for 12 epochs with a step scheduler at epoch 10, taking approximately 18 hours to train on four A100 GPUs with a batch size of 1 for each GPU.

\subsection{Comparison to the state-of-the-art}
\label{sec:experiments:main_results}

\begin{figure*}
\begin{center}
\includegraphics[width=\linewidth]{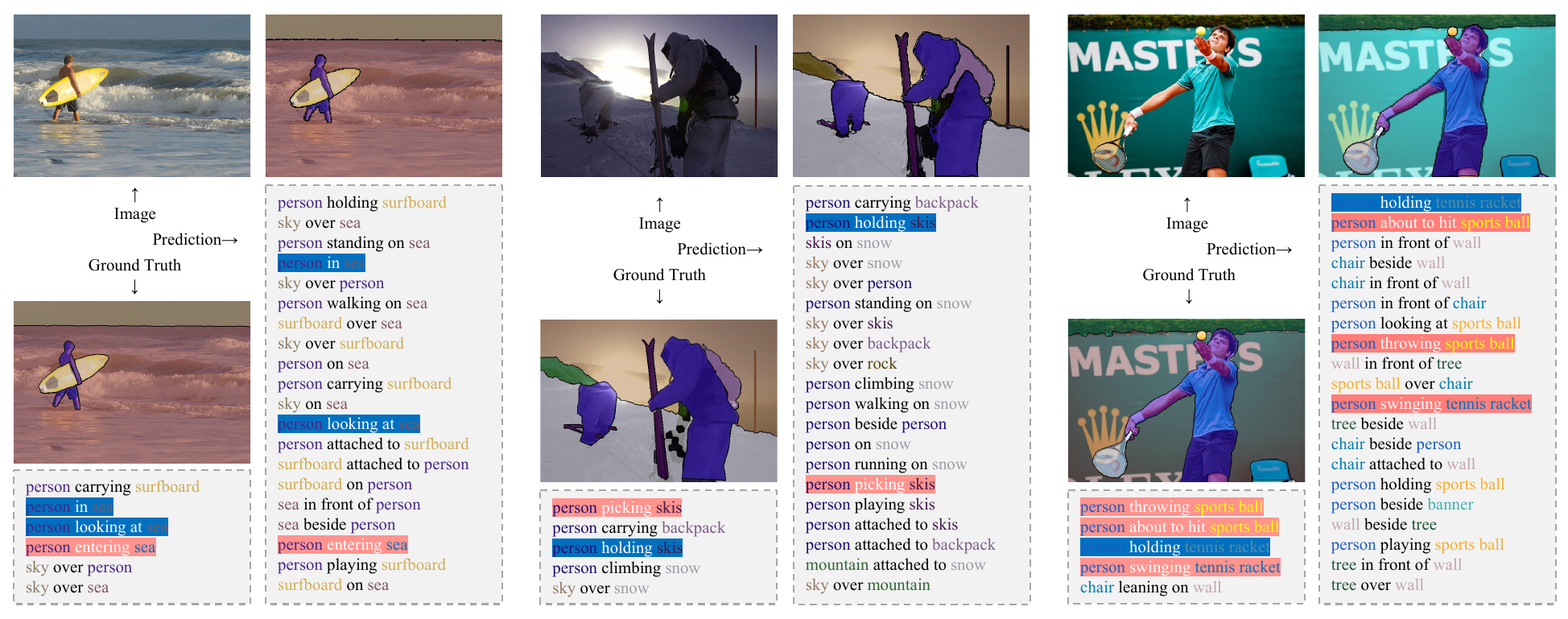}
\end{center}
    \vspace{-5mm}
   \caption{Visualization of panoptic segmentations and the top 20 predicted triplets compared with ground truth.
   The upper left is the original image, the lower left is the ground truth and on the right are the predictions.
   The highlighted triplets represent the subject-object pairs with multiple relations, where the blue highlights represent the high frequency relations and the red highlights represents the low frequency relations.
   The visualization results show that our method can predict both high frequency and low frequency relations.
   }
\vspace{-4mm}
\label{fig:visual}
\end{figure*}

\noindent \textbf{PSG}. 
Tab.~\ref{table:psg_main_results} reports the performance of our method compared to the state-of-the-art on the PSG dataset~\cite{yang2022panoptic}.
We separate the methods into two groups. 
The first are two-stage methods consisting of a separate segmentor and relation predictor, which are modified for the PSG task in \cite{yang2022panoptic}.
The second are one-stage end-to-end methods, which are able to simultaneously predict panoptic segmentation and relations.
Our method belongs to the second category.
For a fair comparison between the different methods, we use the same Resnet-50 \cite{he2016deep} backbone.
Our method shows superior performance compared to all previous methods.
Particularly, it outperforms the previous best-performing method PSGTR \cite{yang2022panoptic} by a large margin, \ie +6\% in R@100 and +11\% in mR@100.
Our model is able to converge within only 12 epochs of training, whereas PSGTR \cite{yang2022panoptic} is trained for 60 epochs.
We also evaluate our method using pre-trained transformer-based backbones, \ie Swin-B and Swin-L~\cite{liu2021swin}, with the latter being a bigger model.
Our results are consistently improved over all metrics due the powerful feature representation ability of the transformer.
We also conducted a visual comparison, as shown in Fig. \ref{fig:visual}.

\noindent \textbf{SGG}.
Here we study whether our approach that was developed for the PSG task can also be applied to the SGG task.
In Tab.~\ref{table:vg_main_results}, we conduct experiments on the VG-150 dataset.
Following IETrans~\cite{zhang2022fine}, we extend our \emph{unbiased} HiLo framework (Sec.~\ref{sec:framework}) without our PSG-specific baseline (Sec.~\ref{sec:baseline}) to four state-of-the-art \emph{biased} SGG methods, namely MOTIF \cite{zellers2018neural}, VCTree \cite{tang2019learning}, Transformer \cite{tang2020unbiased} and GPSNet \cite{lin2020gps}, using the implementation of \cite{zhang2022fine}.
Compared to the unbiased IETrans \cite{zhang2022fine} method, our method improves mean recall without sacrificing recall, demonstrating its effectiveness in enhancing the performance of low-frequency relations.
Compared to the biased baselines \cite{zellers2018neural, tang2019learning, tang2020unbiased, lin2020gps}, our method achieves a significantly larger mean recall, while still maintaining an acceptable recall.
This indicates that our method can improve the performance of low-frequency relations while also taking into account the performance of high-frequency relations.
It also shows that the HiLo framework is a general technique that yields systematic improvements in both the PSG and SGG tasks.

\begin{table}\small
    \centering
    \begin{tabular}{l|cc}
    \hline
        ~  & \multicolumn{2}{c}{Scene Graph Detection} \\
        \cline{2-3}
        Method & R/mR@50 & R/mR@100 \\
        \hline
        MOTIF \cite{zellers2018neural} & \textbf{31.0} / 6.7 & \textbf{35.1} / 7.7 \\
        \quad +IETrans \cite{zhang2022fine} & 26.4 / 12.4 & 30.6 / 14.9 \\
        \quad \textbf{+HiLo (ours)} & 26.2 / \textbf{14.7} & 30.3 / \textbf{17.7}  \\
        \hline
        VCTree \cite{tang2019learning} & \textbf{30.2} / 6.7 & \textbf{34.6} / 8.0 \\
        \quad +IETrans \cite{zhang2022fine} & 25.4 / 11.5 & 29.3 / 14.0 \\
        \quad \textbf{+HiLo (ours)} & 27.1 / \textbf{12.9} & 29.8 / \textbf{15.2} \\
        \hline
        Transformer \cite{tang2020unbiased} & \textbf{30.0} / 7.4 & \textbf{34.3} / 8.8 \\
        \quad +IETrans \cite{zhang2022fine} & 25.5 / 12.5 & 29.6 / 15.0 \\
        \quad \textbf{+HiLo (ours)} & 25.4 / \textbf{14.6} & 29.2 / \textbf{17.6} \\
        \hline
        GPSNet \cite{lin2020gps} & \textbf{30.3} / 5.9 & \textbf{35.0} / 7.1 \\
        \quad +IETrans \cite{zhang2022fine} & 25.9 / 14.6 & 28.1 / 16.5 \\
        \quad \textbf{+HiLo (ours)} & 25.6 / \textbf{15.8} & 27.9 / \textbf{18.0} \\
        \hline
    \end{tabular}
    \vspace{+1mm}
    \caption{Comparison between our HiLo framework and other methods on the VG-150 dataset. Similar to \cite{zhang2022fine}, we apply IETrans and our own method on top of four leading baselines.
    }
    \vspace{-5mm}
    \label{table:vg_main_results}
\end{table}

\subsection{Ablation Studies}
\label{sec:experiments:ablation}
Consistent with the paper, we use HiLo with a Resnet-50 backbone to perform ablation experiments on the PSG dataset.

\noindent \textbf{HiLo framework for different baselines}.
In this section we investigate whether our HiLo framework~(Sec.~\ref{sec:framework}) yields improvements for other baselines, rather than just the one presented in Sec.~\ref{sec:baseline}.
Tab.~\ref{table:different_baselines} shows the results for two baselines, with and without the HiLo framework.
We observe that our biased baseline outperforms the previous PSGTR~\cite{yang2022panoptic} method on all metrics.
Furthermore, by applying the HiLo framework, we can substantially improve the performance over both baselines.
It is worth mentioning that the HiLo framework improves recall and mean recall simultaneously, whereas other methods typically improve one metric at the cost of the other \cite{tang2020unbiased, yu2020cogtree}.

\begin{table}\small
    \centering
    \resizebox{\linewidth}{!}{
    \begin{tabular}{lc|ccc}
    \hline
        Baseline  & HiLo & R/mR@20 & R/mR@50 & R/mR@100 \\
        \hline
        HiLo baseline & \checkmark & \textbf{34.1} / \textbf{23.7} & \textbf{40.7} / \textbf{30.3} & \textbf{43.0} / \textbf{33.1} \\
        HiLo baseline & - & 32.6 / 20.9 & 38.0 / 27.4 & 38.9 / 28.4 \\
        \hline
        PSGTR~\cite{yang2022panoptic} & \checkmark & \textbf{30.1} / \textbf{20.2} & \textbf{36.6} / \textbf{23.9} & \textbf{38.3} / \textbf{24.5} \\
        PSGTR~\cite{yang2022panoptic} & - & 28.4 / 16.6 & 34.4 / 20.8 & 36.3 / 22.1 \\
        \hline
    \end{tabular}
    }
    \vspace{+1mm}
    \caption{Comparison of different baselines, with and without HiLo framework. Using the HiLo framework, we see significant improvements on both metrics.
    }
    \vspace{-2mm}
    \label{table:different_baselines}
\end{table}

\noindent \textbf{HiLo relation augmentation.}
We observe that out of 260,296 labeled triplets in the PSG dataset, only about 10\% of subject-object pairs have multiple relations, for which we can apply relation swapping~(Sec.~\ref{sec:framework:relation_generation}).
After applying our proposed relation augmentation technique~(Sec.~\ref{sec:framework:relation_generation}), this ratio significantly increases to 40\%.
Our experimental results in Tab.~\ref{table:proportion_of_swappable_triplets} demonstrate that only applying the HiLo framework on 10\% already gives an improvement over the baseline from Tab.~\ref{table:different_baselines}. As the number of swappable triplets increases due to augmentation, the model's performance is further enhanced, highlighting the potential of our method.

\begin{table}\small
    \centering
    \resizebox{\linewidth}{!}{
    \begin{tabular}{c|c|ccc}
    \hline
        Relation Aug. & Multiple relations & R/mR@50 & R/mR@100 \\
        \hline
        \checkmark & 40\%  & \textbf{40.7} / \textbf{30.3} & \textbf{43.0} / \textbf{33.1} \\
        - & 10\%  & 40.1 / 28.1 & 42.8 / 32.5 \\
        \hline
    \end{tabular}
    }
    \vspace{+1mm}
    \caption{Ablation study for HiLo relation augmentation. Relation augmentation affects the ratio of subject-object pairs with multiple relations. The larger this ratio, the more relations can be swapped, which leads to better results.
    }
    \vspace{-4mm}
    \label{table:proportion_of_swappable_triplets}
\end{table}


\noindent \textbf{HiLo prediction alignment}.
We conduct ablation experiments on the subject-object consistency loss and relation consistency loss~(Sec.~\ref{sec:framework:prediction_alignment}), which are used to align the predictions from the high and low frequency branches.
The results, as presented in Tab.~\ref{table:prediction_alignment}, demonstrate that using both losses yields the best performance.
It is worth mentioning that we have explored the margin in the relation consistency loss and found that setting the margin to zero leads to a small performance degradation.
This finding confirms that there is a partial semantic overlap between swapped relations, indicating that they are not entirely consistent.

To investigate the impact of relational index exchange (RIE) on the relation consistency loss, we conducted experiments to verify the effect of omitting RIE.
In the absence of RIE, we solely compute the consistency loss for relation categories that are not involved in relation swapping and exclude the swapping component from the calculation of the consistency loss.
The outcomes of this experiment are presented in Tab. \ref{table:rie}, and demonstrate a notable reduction in the mean recall and a decline in the model's performance for relations with relational semantic overlap when RIE is not utilized.

\begin{table}\small
    \centering
    \resizebox{\linewidth}{!}{
    \begin{tabular}{ccc|ccc}
    \hline
        Object & Relation & Margin & R/mR@50 & R/mR@100 \\
        \hline
        \checkmark & \checkmark & 0.5 & \textbf{40.7} / \textbf{30.3} & \textbf{43.0} / \textbf{33.1} \\
        - & \checkmark & 0.5 & 40.6 / 29.7 & 42.8 / 32.8 \\
        - & \checkmark & 0.0 &  40.5 / 29.5 & 42.7 / 32.8 \\
        \checkmark & - & - &  40.4 / 29.0 & 42.8 / 32.2 \\
        - & - & - & 39.7 / 28.6 & 42.4 / 32.0 \\
        \hline
    \end{tabular}
    }
    \vspace{+1mm}
    \caption{Ablation study for different losses in HiLo prediction alignment. \emph{Object} refers to the subject-object consistency loss and \emph{relation} refers to the relation consistency loss. The margin parameter is defined in Eq.~\ref{eq:relation_consistency_loss}.
    }
    \vspace{-4mm}
    \label{table:prediction_alignment}
\end{table}

\begin{table}\small
    \centering
    \resizebox{\linewidth}{!}{
    \begin{tabular}{c|ccc}
    \hline
        Whether to use RIE & R/mR@20 & R/mR@50 & R/mR@100 \\
        \hline
        \checkmark & \textbf{34.1} / \textbf{23.7} & \textbf{40.7} / \textbf{30.3} & \textbf{43.0} / \textbf{33.1} \\
        - & 33.5 / 22.3 & 40.3 / 29.0 & 42.6 / 32.3 \\
        \hline
    \end{tabular}
    }
    \vspace{+1mm}
    \caption{Ablation study for relation consistency loss in HiLo prediction alignment.}
    \vspace{-6mm}
    \label{table:rie}
\end{table}

\noindent \textbf{HiLo inference fusion}.
We ablate the inference fusion~(Sec.~\ref{sec:framework:inference_fusion}) and evaluate the performance of each branch's output separately.
In Tab.~\ref{table:inference_fusion_1}, experimental results suggest that fusion can effectively leverage the uniqueness of the high and low frequency branch predictions to achieve comprehensive improvements on all metrics.

We also attempted to average the tensor generated by the two branches and obtain the PSG result through post-processing.
However, we found that this approach leads to a substantial drop in performance, as evident in Tab. \ref{table:inference_fusion_2}.
This can be attributed to the inconsistent prediction results of the two branches for the same query index.
These findings validate that the inference fusion method effectively merges the results from the two branches.
Furthermore, our experimental results demonstrate that the query associated with the identical index in two branches does not predict the same subject-object pair.
Thus, directly averaging the tensor produced by the two branches results in prediction ambiguity, ultimately leading to a substantial decline in performance.
This observation underscores the necessity of conducting triplet query correspondence when performing prediction alignment.
In particular, due to the inconsistent query prediction content for the corresponding index in the two branches, a one-to-one correspondence must be constructed based on the label assigned by each query to achieve prediction alignment.

\begin{table}\small
    \centering
    \resizebox{\linewidth}{!}{
    \begin{tabular}{cc|ccc}
    \hline
        H-L Result & L-H Result & R/mR@50 & R/mR@100 \\
        \hline
        \checkmark & \checkmark & \textbf{40.7} / \textbf{30.3} & \textbf{43.0} / \textbf{33.1} \\
        \checkmark & - & 38.8 / 29.9 & 39.8 / 30.9 \\
        - & \checkmark & 38.5 / 26.5 & 39.5 / 27.8 \\
        \hline
    \end{tabular}
    }
    \vspace{+1mm}
    \caption{Ablation study for HiLo inference fusion.}
    \vspace{-4mm}
    \label{table:inference_fusion_1}
\end{table}

\begin{table}\small
    \centering
    \resizebox{\linewidth}{!}{
    \begin{tabular}{l|ccc}
    \hline
        Fusion method & R/mR@20 & R/mR@50 & R/mR@100 \\
        \hline
        inference fusion & \textbf{34.1} / \textbf{23.7} & \textbf{40.7} / \textbf{30.3} & \textbf{43.0} / \textbf{33.1} \\
        average tensor & 19.6 / 13.1 & 23.1 / 15.7 & 23.9 / 16.3 \\
        \hline
    \end{tabular}
    }
    \vspace{+1mm}
    \caption{Ablation study for HiLo inference fusion.
    \textit{Inference fusion} refers to the method proposed in the paper to fuse the results of two branches.
    \textit{Average tensor} refers to the fusion method that directly averages the tensor output by the two branches.
    }
    \vspace{-4mm}
    \label{table:inference_fusion_2}
\end{table}

\noindent \textbf{Masked relation attention}.
We investigate the impact of different mask input types for cross-attention on the HiLo baseline~(Sec.~\ref{sec:baseline}) performance.
Specifically, we compare two different attention focus regions, namely the subject-object region and the full image.
The results are shown in Tab.~\ref{table:masked_relation_attention}.
Focusing on the full image presents a more challenging optimization task for the model since no target region is specified.
Consequently, we observe a drop of 1.8\% in R@100 and 1.6\% in mR@100.
This shows that it is crucial to apply masked relation attention to the subject and object.

\begin{table}\small
    \centering
    \resizebox{\linewidth}{!}{
    \begin{tabular}{l|ccc}
    \hline
        Attention focus & R/mR@20 & R/mR@50 & R/mR@100 \\
        \hline
        subject-object & \textbf{32.6} / \textbf{20.9} & \textbf{38.0} / \textbf{27.4} & \textbf{38.9} / \textbf{28.4} \\
        full image & 30.4 / 19.3 & 36.5 / 25.9 & 37.1 / 26.8 \\
        \hline
    \end{tabular}
    }
    \vspace{+1mm}
    \caption{Ablation study for masked relation attention.}
    \vspace{-10mm}
    \label{table:masked_relation_attention}
\end{table}

\subsection{Analysis}
\label{exp:analysis}
To further demonstrate the efficacy of our method, we conduct the following analysis on HiLo.

\noindent \textbf{Long-tail problem and relational semantic overlap.}
To verify whether the problems of long-tail and relational semantic overlap have been tackled, we conduct experiments shown in Tab. \ref{table:verify}.
For the \textit{long-tail problem}, we consider relations appearing less than 500 times (28 relations) in the PSG dataset (relations appear on average 4787 times) as rare relations, and report their mR@100.
There is a 14\% improvement for rare relations and an 11\% improvement for all relations.
This suggests that our method addresses the long-tail problem.

For the \textit{relational semantic overlap}, we select all test images that have this problem (927 images) and report their mR@100.
Our method shows a 15.3\% improvement over PSGTR on images with semantic overlap relations, exceeding the overall improvement across all test images.
This suggests that our approach addresses the problem of relational semantic overlap.
\begin{table}\small
    \centering
    \resizebox{\linewidth}{!}{
    \begin{tabular}{l|cc|cc}
    \hline
        Method & All & Rare & All & Overlap \\ \hline
        PSGTR & 22.1 & 6.2 & 22.1 & 23.5 \\
        HiLo (ours) & 33.1 (\textcolor{darkgreen}{+11.0}) & 20.3 (\textcolor{darkgreen}{+14.1}) & 33.1 (\textcolor{darkgreen}{+11.0}) & 38.8 (\textcolor{darkgreen}{+15.3}) \\ \hline
    \end{tabular}
    }
    \vspace{+1mm}
    \caption{Verification of the improvements in the long-tail problem and relational semantic overlap.
    \textit{All} refers to testing on the whole test set.
    \textit{Rare} refers to testing only on rare relations.
    \textit{Overlap} refers to testing only on data with relation semantic overlap.}
    \vspace{-4mm}
    \label{table:verify}
\end{table}



\begin{figure}
\begin{center}
\includegraphics[width=\columnwidth]{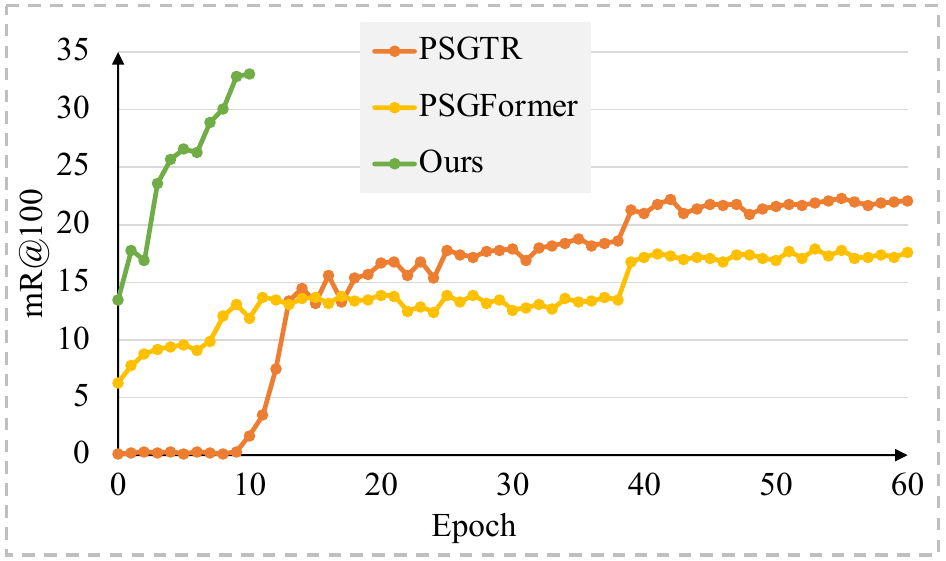}
\end{center}
\vspace{-2mm}
\caption{Convergence speed analysis of different methods. 
Our method converges significantly faster than previous methods.}
\vspace{-6mm}
\label{figure:convergence_speed}
\end{figure}

\noindent \textbf{Convergence speed and time cost analysis}.
We evaluate the convergence of our model by assessing its performance on the validation set at various epochs, as illustrated in Fig.~\ref{figure:convergence_speed}.
Our analysis reveals that our proposed method outperforms prior methods \cite{yang2022panoptic} both in terms of final performance and convergence speed.
Specifically, PSGTR \cite{yang2022panoptic} achieves negligible performance in the initial 12 epochs, requiring 60 epochs to converge, as per the authors.
In contrast, our HiLo method achieves better results in just 12 epochs, indicating its superior convergence speed.

For time cost, we primarily analyze relation augmentation and swapping.
1) \textit{Relation augmentation}, inspired by IETrans \cite{zhang2022fine}, involves training a baseline model and then using it to predict relation labels so as to augment the original relation labels. 
For the PSG dataset, following our experimental setup (see Sec. \ref{sec:experiments:implementation}), this takes 18 hours.
Afterwards, the training of our HiLo model requires 18 hours, which makes the whole process 36 hours.
In contrast, the PSGTR model training takes 48 hours.
Both our method and PSGTR utilize ResNet-50 as the backbone for fair comparison.
Our method's rapid convergence (see Sec. \ref{sec:experiments:ablation}) reduces training time compared to PSGTR, making the additional time cost for relation augmentation tolerable.
2) \textit{Relation swapping} is a quick operation on relation labels during training and does not significantly contribute to overall time consumption.

\noindent \textbf{Training and inference cost.}
Training and inference cost is shown in Tab. \ref{table:resource}.
Despite using two transformer-based decoders, our method requires less resource.
Given the same input sizes $(1280, 800)$ and ResNet-50 as backbone, the resource usage is shown in Tab. \ref{table:resource}.
PSGTR \cite{yang2022panoptic} generates mask features for subject and object separately, while our method reduces this computation by only generating mask features once.
Our inference time is marginally higher due to more complex post-processing.
Using H-L and L-H data only changes the labels between two branches, not increasing resource use.
\begin{table}\small
    \centering
    \resizebox{\linewidth}{!}{
    \begin{tabular}{l|cccc}
    \hline
        Method & GFlops & Param. (M) & Train Mem. (G) & Infer. Time (ms)\\ \hline
        PSGTR & 461.3 & 44.2 & 26.5 & 140 \\
        HiLo (ours) & 229.4 & 58.7 & 16.1 & 156 \\ \hline
    \end{tabular}
    }
    \vspace{+1mm}
    \caption{Training and inference cost.}
    \label{table:resource}
    \vspace{-6mm}
\end{table}

%% file: 5_conclusion.tex
In this work we proposed the HiLo framework to tackle the long-tail problem with relational semantic overlap in Panoptic Scene Graph generation.
The HiLo framework simultaneously learns the high and low frequency relations in different network branches and unifies their strengths by aligning their predictions.
We also constructed a HiLo baseline to allow high-quality panoptic segmentation to improve PSG performance.
Experimental results demonstrate that our method achieves state-of-the-art performance on the PSG dataset, confirming its effectiveness.
In future work, we will investigate how knowledge distillation \cite{hinton2015distilling, gou2021knowledge} can be used to fuse the high and low branches in our method, as well as its application to downstream tasks such as visual question answering and image captioning.

%% file: 6_appendix.tex
\appendix

To further substantiate the effectiveness of our proposed method, we provide additional experimental results in the supplementary material that could not be included in the paper due to space limitations.
The supplementary material includes more experimental results on VG-150.

\section{More results on VG-150}
We present an extension of our previous results on scene graph detection (SGDet), the most challenging task in the VG-150 \cite{krishna2017visual} dataset, by including comparative evaluations on two simplified tasks: predicate (relation) classification (PredCls) and scene graph classification (SGCls).
PredCls involves inferring the relations between objects while assuming prior knowledge of their location and category information.
On the other hand, SGCls entails the prediction of the category of an object and the relations between objects, given only their location information.
We provide the detailed results of these tasks in the Tab. \ref{table:vg_main_results_predcls} and Tab. \ref{table:vg_main_results_sgcls}.

These results once again demonstrate that our method can improve the performance of low-frequency relations while also taking into account the performance of high-frequency relations.
It also shows that the HiLo framework is a general technique that yields systematic improvements in both the panoptic scene graph generation (PSG) and scene graph generation (SGG) tasks.


\begin{table}[h!]\small
    \centering
    \begin{tabular}{l|cc}
    \hline
        ~  & \multicolumn{2}{c}{Predicate Classification} \\
        \cline{2-3}
        Method & R/mR@50 & R/mR@100 \\
        \hline
        MOTIF \cite{zellers2018neural} & \textbf{64.0} / 15.2 & \textbf{66.0} / 16.2 \\
        \quad +IETrans \cite{zhang2022fine} & 54.7 / 30.9 & 56.7 / 33.6 \\
        \quad \textbf{+HiLo (ours)} & 53.6 / \textbf{33.6} & 55.5 / \textbf{36.4}  \\
        \hline
        VCTree \cite{tang2019learning} & \textbf{64.5} / 16.3 & \textbf{66.5} / 17.7 \\
        \quad +IETrans \cite{zhang2022fine} & 53.0 / 30.3 & 55.0 / 33.9 \\
        \quad \textbf{+HiLo (ours)} & 53.4 / \textbf{34.0} & 55.2 / \textbf{37.8} \\
        \hline
        Transformer \cite{tang2020unbiased} & \textbf{63.6} / 17.9 & \textbf{65.7} / 19.6 \\
        \quad +IETrans \cite{zhang2022fine} & 51.8 / 30.8 & 53.8 / 34.7 \\
        \quad \textbf{+HiLo (ours)} & 52.9 / \textbf{32.8} & 55.9 / \textbf{36.1} \\
        \hline
        GPSNet \cite{lin2020gps} & \textbf{65.1} / 15.0 & \textbf{66.9} / 16.0 \\
        \quad +IETrans \cite{zhang2022fine} & 52.3 / 31.0 & 54.3 / 34.5 \\
        \quad \textbf{+HiLo (ours)} & 53.3 / \textbf{33.8} & 55.2 / \textbf{37.4} \\
        \hline
    \end{tabular}
    \vspace{+1mm}
    \caption{Comparison between our HiLo framework and other methods on PredCls on the VG-150 dataset. Similar to \cite{zhang2022fine}, we apply IETrans and our own method on top of four leading baselines.
    }
    \label{table:vg_main_results_predcls}
\end{table}

\begin{table}[h!]\small
    \centering
    \begin{tabular}{l|cc}
    \hline
        ~  & \multicolumn{2}{c}{Scene Graph Classification} \\
        \cline{2-3}
        Method & R/mR@50 & R/mR@100 \\
        \hline
        MOTIF \cite{zellers2018neural} & \textbf{38.0} / 8.7 & \textbf{38.9} / 9.3 \\
        \quad +IETrans \cite{zhang2022fine} & 32.5 / 16.8 & 33.4 / 17.9 \\
        \quad \textbf{+HiLo (ours)} & 32.1 / \textbf{18.9} & 33.1 / \textbf{20.9}  \\
        \hline
        VCTree \cite{tang2019learning} & \textbf{39.3} / 8.9 & \textbf{40.2} / 9.5 \\
        \quad +IETrans \cite{zhang2022fine} & 32.9 / 16.5 & 33.8 / 18.1 \\
        \quad \textbf{+HiLo (ours)} & 35.7 / \textbf{21.0} & 36.8 / \textbf{22.7} \\
        \hline
        Transformer \cite{tang2020unbiased} & \textbf{38.1} / 9.9 & \textbf{39.2} / 10.5 \\
        \quad +IETrans \cite{zhang2022fine} & 32.6 / 17.4 & 33.5 / 19.1 \\
        \quad \textbf{+HiLo (ours)} & 32.3 / \textbf{20.1} & 33.3 / \textbf{22.2} \\
        \hline
        GPSNet \cite{lin2020gps} & \textbf{36.9} / 8.2 & \textbf{38.0} / 8.7 \\
        \quad +IETrans \cite{zhang2022fine} & 31.8 / 17.0 & 32.7 / 18.3 \\
        \quad \textbf{+HiLo (ours)} & 31.7 / \textbf{18.3} & 32.5 / \textbf{20.2} \\
        \hline
    \end{tabular}
    \vspace{+1mm}
    \caption{Comparison between our HiLo framework and other methods on SGCls on the VG-150 dataset. Similar to \cite{zhang2022fine}, we apply IETrans and our own method on top of four leading baselines.}
    \vspace{+4mm}
    \label{table:vg_main_results_sgcls}
\end{table}